**ARTICLE TYPE**

# Skip-WaveNet: A Wavelet based Multi-scale Architecture to Trace Snow Layers in Radar Echograms

Debvrat Varshney[1], Masoud Yari[2], Oluwanisola Ibikunle[3], Jilu Li[3], John Paden[3], Aryya Gangopadhyay[1] and Maryam Rahnemoonfar[2,4]

[1]Department of Information Systems, University of Maryland Baltimore County, Baltimore, 21250, Maryland, USA.
[2]Department of Computer Science and Engineering, Lehigh Univerisity, Bethlehem, 18015, Pennsylvania, USA.
[3]Center for Remote Sensing and Integrated Systems, University of Kansas, Lawrence, 66045, Kansas, USA.
[4]Department of Civil and Environmental Engineering, Lehigh Univerisity, Bethlehem, 18015, Pennsylvania, USA.
E-mail: maryam@lehigh.edu.

**Abstract**
Airborne radar sensors capture the profile of snow layers present on top of an ice sheet. Accurate tracking of these layers is essential to calculate their thicknesses, which are required to investigate the contribution of polar ice cap melt to sea-level rise. However, automatically processing the radar echograms to detect the underlying snow layers is a challenging problem. In our work, we develop wavelet-based multi-scale deep learning architectures for these radar echograms to improve snow layer detection. These architectures estimate the layer depths with a mean absolute error of 3.31 pixels and 94.3% average precision, achieving higher generalizability as compared to state-of-the-art snow layer detection networks. These depth estimates also agree well with physically drilled stake measurements. Such robust architectures can be used on echograms from future missions to efficiently trace snow layers, estimate their individual thicknesses and thus support sea-level rise projection models.

## Impact Statement

Global warming is a reality that is harshly affecting the polar ice caps. The increase in their annual melt rates and their contribution to global sea-level rise can potentially cause drastic socio-economic damage. Radar devices are the most popular sensors to monitor ice caps, but processing and analyzing their data is not straightforward. In this work, we propose a deep learning algorithm that can effectively process polar radar echograms and calculate the thickness of snow accumulated on top of an ice sheet. The estimated thicknesses can be used by glaciological models to project global sea-level rise, eventually helping us prepare for any future calamities.

## 1. Introduction

Increase of global mean annual temperature every year is having a drastic effect on the polar ice-caps. Intergovernmental Panel on Climate Change (IPCC) 2021 report (Masson-Delmotte et al., 2021) states that the Arctic has warmed at more than twice the global rate over the past 50 years, and the melting of the Greenland Ice Sheet (GrIS) will cause sea levels to rise by 2 meters by 2100, potentially flooding regional coastal areas (Kirezci et al., 2020). Modern climate models that simulate and project sea-level rise rely on the thickness of the snow accumulated on top of the ice caps (Koenig et al., 2016). Continuous monitoring and evaluation of this thickness is hence imperative to support climate models make accurate projections and prepare us for future natural disasters.

Polar ice sheets are thick masses of ice, on top of which snow gets accumulated every year to form snow accumulation layers. The change in volume of these snow layers changes the surface mass balance (SMB) of the ice sheet.If the SMB of an ice-sheet turns negative, it results in sea-level increase. Airborne radar systems such as the Snow Radar (Gogineni et al., 2013) monitor and capture the internal state of these snow layers annually. Significant variations in snow permittivity reflect the electromagnetic signals transmitted by Snow Radar. This reflection, which can be seen as the layers in radar echograms, often occurs at the boundaries of snow stratigraphy, where changes in the physical properties of the snow, such as density, hardness, grain size, and shape, result in significant discontinuities in the snow's dielectric properties. However, the visibility and clarity of the snow layers in Snow Radar echograms degrades when the magnitude of the discontinuity in the snow's dielectric properties decreases. Routine data processing has applied deconvolution, filtering, coherent and incoherent integration techniques to remove non-ideal system characteristics and improve the signal-to-noise ratio; however, off-nadir surface backscattering, multipath scattering, and signal loss in the medium may still exacerbate the degradation of Snow Radar echogram's quality (see the examples in the leftmost columns of Figures 8-10).

Due to the degraded quality of the echograms, tracking snow accumulation layers is challenging for experienced glaciologists as well as conventional vision algorithms (Koenig et al., 2016; Rahnemoonfar et al., 2021). Even convolutional neural networks (CNNs) (LeCun et al., 2015), which have become the standard machine learning algorithms for image processing, have issues in making accurate predictions from these images (Rahnemoonfar et al., 2021; Varshney, Rahnemoonfar, Yari, Paden, et al., 2021; Varshney et al., 2020) since any interruptions to their input can drastically affect their prediction capability (Q. Li et al., 2021). To improve the effectiveness of CNNs in processing these radar images, we use wavelet transforms (Mallat, 1989), which are signal processing techniques that can represent an image in a multi-resolution format. They have localization properties, in both the spatial and frequency domains, and can depict the contextual as well as textural information in the image at different scales through their detail coefficients (Huang et al., 2017; Mallat, 1989). We use these detail coefficients to guide the CNN towards an enhanced snow layer representation.

In this work, we exploit the multi-scale nature of wavelet transforms and fuse them in a multi-scale convolutional neural network (Rahnemoonfar et al., 2021) to improve the accuracy of snow layer detection. We also show that taking wavelet transforms of intermediate scales of the neural network helps in learning and extracting features, as compared to taking multi-level wavelet transform of the input image. This work is an extension of Varshney, Yari, et al. (2021) with a stronger backbone network, more types of wavelet transforms, analysis of 'static' wavelets vs 'dynamic' wavelets, comparison with state-of-the-art networks, calculation of individual layer depths (in terms of number of pixels), comparison with in-situ stake measurements, along with the use of a more practical Snow Radar dataset.

The rest of the paper is organized as follows - we highlight the past work on snow layer tracking and wavelet combined neural networks in Section 2; give a brief background of wavelet transforms, explain the proposed wavelet based multi-scale architecture(s), and showcase the evaluation techniques in Section 3; describe the Snow Radar data and the in-situ observations that we use for our experiments in Section 4; discuss the quantitative and qualitative results in Section 5, and finally conclude the paper in Section 6 highlighting some future work.

## 2. Related Works

Recently, there have been extensive studies on using deep learning for tracking snow layers through radar echograms. At the same time, wavelet transformations are being exhaustively used to improve the feature extraction capabilities of deep learning networks. Here, we provide a quick overview of the research being done in these two areas.

### 2.1. *Snow Layer Tracking through Deep Learning*

Feature extraction capabilities and good generalizability of deep learning networks have been used on Snow Radar images to track snow layers in Yari et al. (2019), Yari et al. (2020), Rahnemoonfar et al. (2021), Varshney, Rahnemoonfar, Yari, Paden, et al. (2021), Varshney, Yari, et al. (2021), and Wang et al. (2021). Deep learning has also been used to segment the radar depth sounder (RDS) images in Donini et al. (2022), Ghosh and Bovolo (2022a), and Ghosh and Bovolo (2022b) into ice and bedrock through various modifications of UNet (Ronneberger et al., 2015), with or without transformers (Vaswani et al., 2017). In this paper, we focus on the shallow sensor, the Snow Radar (Gogineni et al., 2013), which track the snow layers.

Wang et al. (2021) used a CNN-RNN (recurrent neural network) architecture for tiered segmentation of radar images to identify the top 4-6 snow layers. Similarly, Ibikunle et al. (2020) built an iterative neural network architecture to track snow layers, but on simulated radar images, as real radar images were too complex for feature extraction. Varshney, Rahnemoonfar, Yari, Paden, et al. (2021) used pyramid pooling modules to learn local-to-global spatio-contextual information of snow layer pixels and perform semantic segmentation. In the line of multi-scale architectures, Yari et al. (2019) first explored multi-scale contour detection CNNs for the purpose of snow layer extraction. The authors noted that using popular pre-training strategies would not work on Snow Radar images due to the inherent noise present in these images. Further extensions of this work by Rahnemoonfar et al. (2021) and Yari et al. (2020) showed the usefulness of multi-scale architectures on synthetic radar images and temporal transfer learning, respectively. Subsequently, wavelet based multi-scale architectures for snow layer tracking were first developed by Varshney, Yari, et al. (2021) which showed preliminary results in this domain. This prior work used a VGG-13 architecture (Simonyan & Zisserman, 2015) on a small Snow Radar dataset, which achieved an F-score of little over 0.7. We expand upon this work by training a VGG-16 architecture, which is known to give robust representations (Rahnemoonfar et al., 2021), on a new dataset specifically catered towards dry zone areas, where the snow layers can be tracked. Further, and most importantly, we also calculate the individual layer depths in terms of number of pixels. We also show that using wavelet transforms of each scale of a multi-scale network helps in learning inherent image features, as compared to using multi-level wavelet transforms of the input image.

### 2.2. *CNNs with Wavelet Transformations*

Wavelet transforms have had immense applications in image denoising (Bnou et al., 2020; P. Liu et al., 2019; Mallat, 1989), image compression (Naveen Kumar et al., 2019; Xiong & Ramchandran, 2009), and image restoration (Bae et al., 2017; Huang et al., 2017; P. Liu et al., 2019). Many recent works such as P. Liu et al. (2019), Williams and Li (2018), Huang et al. (2017), Han and Ye (2018), and Bae et al. (2017) have used the downsampling properties of wavelets and replaced the convolutional or pooling layers in a CNN, since incorporating the wavelet coefficients typically helps reduce information loss (P. Liu et al., 2019). Williams and Li (2018) showed that using wavelet transform for downsampling would create much cleaner and sharper images, as compared to using pooling layers, and improve generalizability. Similarly, P. Liu et al. (2019) found that wavelet transforms not only enlarge a kernel's receptive field, but also prevent information leakage which generally takes place during a pooling operation. By using multi-level wavelet transforms, Han and Ye (2018) developed deep convolutional framelets to reconstruct sparse-view computed tomography images and Huang et al. (2017) developed a wavelet based loss function to super-resolve facial images. Further, Bae et al. (2017) showed that residual learning improves by training on wavelet subbands. The variety of these works show that by incorporating wavelet transforms in a CNN, the feature extraction capability of the latter improves due to sharper feature maps, increased receptive field, and enhanced residual learning.

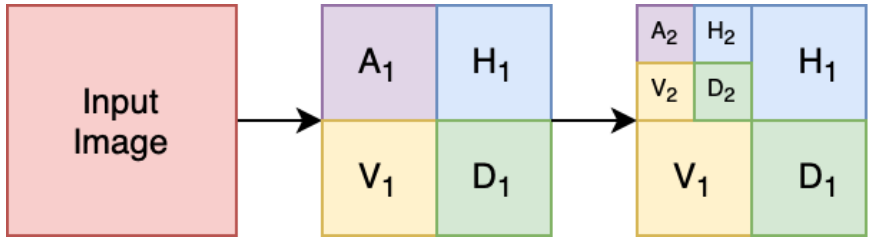

Figure 1: A level 2 wavelet transform of a given input image. The subscript denotes the level number

## 3. Methodology

In this section, we briefly explain the background of wavelet transforms, describe our base architecture without wavelets called Multi-Scale CNN, and then explain the wavelet combined neural networks for snow layer tracking.

### 3.1. Wavelet Transform

We use the Discrete Wavelet Transform (DWT) for our computations. For an image $f(x, y)$ having dimensions $X \times Y$, DWT is given as Equations 3.1 and 3.2 for $j \geq j_0$:

$$W_\phi(j_0, m, n) = \frac{1}{\sqrt{X.Y}} \sum_{x=0}^{X-1} \sum_{y=0}^{Y-1} f(x, y)\phi_{j_0,m,n}(x, y) \tag{3.1}$$

$$W_\psi^i(j, m, n) = \frac{1}{\sqrt{X.Y}} \sum_{x=0}^{X-1} \sum_{y=0}^{Y-1} f(x, y)\psi_{j,m,n}^i(x, y) \tag{3.2}$$

In these equations, $W_\phi$ is the approximation coefficient (A), and $W_\psi^i$ are the detail coefficients for each level of wavelet transform $j$, where $i \in \{H, V, D\}$. H, V, D are the horizontal, vertical, and diagonal details, respectively, $m, n$ are the subband dimensions (Williams, Li, et al., 2018), and $j_0$ is an arbitrary starting level. Each subsequent level of wavelet transform is computed on the previous level's approximation coefficient. In the beginning, the input raw image is treated as as an approximation coefficient. Further, $\phi$ is a scaling function, and $\psi$ is the wavelet function, both of which downsample an input image by a factor of 2 in both X and Y dimentions. Readers are encouraged to go through Daubechies (1992) to get a detailed understanding of $\phi$, $\psi$, and the wavelet basis functions. We use the downsampling property of wavelets and combine their transforms with the side outputs of multi-scale networks. Figure 1 shows a level 2 (i.e. $j = 2$) transform of an input image. In our work, we experiment with three popular discrete wavelets, i.e. Haar, Daubechies, and 'Discrete' Meyer . We will be abbreviating them as 'haar', 'db', and 'dmey' in subsequent sections for ease of reading. The wavelet functions ($\psi$) for the three wavelet types have been plotted in Figure 2.

### 3.2. Multi-Scale CNN (MS-CNN)

In this work, we use a multi-scale network built for contour detection (Xie & Tu, 2015), which was proven to be useful for snow layer tracking (Rahnemoonfar et al., 2021; Yari et al., 2019). This is a VGG-16 (Simonyan & Zisserman, 2015) network with the terminal fully connected layers and the last pooling layer removed. The final convolutional layers of every stage, right before the max pooling layer of every stage, are convolved with a 1×1 filter to generate what we call "side output(s)". Each side output is at a downsampled resolution of the previous stage's side output. All side outputs are finally upsampled through transposed convolutions, followed by cropping, to match the resolution of the input

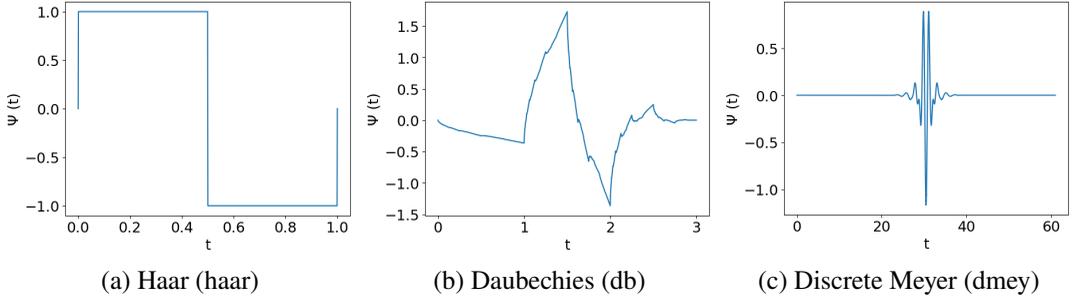

(a) Haar (haar)    (b) Daubechies (db)    (c) Discrete Meyer (dmey)

Figure 2: The three types of wavelets that we use for our experiments, their shorthand representation in parenthesis

image. The five side outputs, from the five stages, are concatenated together to form a 'fuse' layer. All five side outputs along with the final 'fuse' layer are trained together in a deeply supervised manner (Lee et al., 2015) with a cumulative loss function, Equation 3.5, as explained in the following paragraph.

To train the network, we calculate the binary cross entropy loss ($l$) for every pixel $i$ as follows:

$$l(x_i; W) = \begin{cases} \alpha \cdot \log(1 - x_i) & \text{if } y_i = 0 \\ \beta \cdot \log(x_i) & \text{if } y_i = 1 \end{cases} \tag{3.3}$$

In Equation 3.3, $x_i$ is the sigmoid activation map obtained from a network with weights $W$, $y_i$ is the ground truth label of the corresponding pixel in the input image $I$ having a total of $|I|$ pixels, and $\alpha$, $\beta$ are defined as Equation 3.4.

$$\alpha = \lambda \cdot \frac{|Y^+|}{|Y^+| + |Y^-|}$$
$$\beta = \frac{|Y^-|}{|Y^+| + |Y^-|} \tag{3.4}$$

In Equation 3.4, $|Y^+|$ denotes the count of all positive labels, i.e. those pixels representing the top of a layer ($y_i = 1$) and $|Y^-|$ denotes the count of all negative labels i.e. all other pixels which are background ($y_i = 0$). $\lambda$ is a hyperparameter used to balance these positive and negative labels.

The total loss is computed as Equation 3.5 where $k$ depicts each of the side outputs or stages, i.e. $K = 5$. This network, called Multi-Scale CNN, is the current state-of-the-art model for snow layer tracking (Rahnemoonfar et al., 2021) and forms our baseline architecture to compare with our wavelet-based architectures. Multi-Scale CNN is shown in Figure 3 and most of the times abbreviated as MS-CNN for the rest of the paper.

$$L(W) = \sum_{i=1}^{|I|} \left( \sum_{k=1}^{K} l(x_i^k; W) + l(x_i^{fuse}; W) \right) \tag{3.5}$$

### 3.3. Wavelet combined Neural Networks

We set up two wavelet based architectures, and exploit the multi-level nature of wavelet transforms to embed them into the base multi-scale architecture, MS-CNN. The two wavelet based network modifications are termed as 'WaveNet' and 'Skip-WaveNet', and described below:

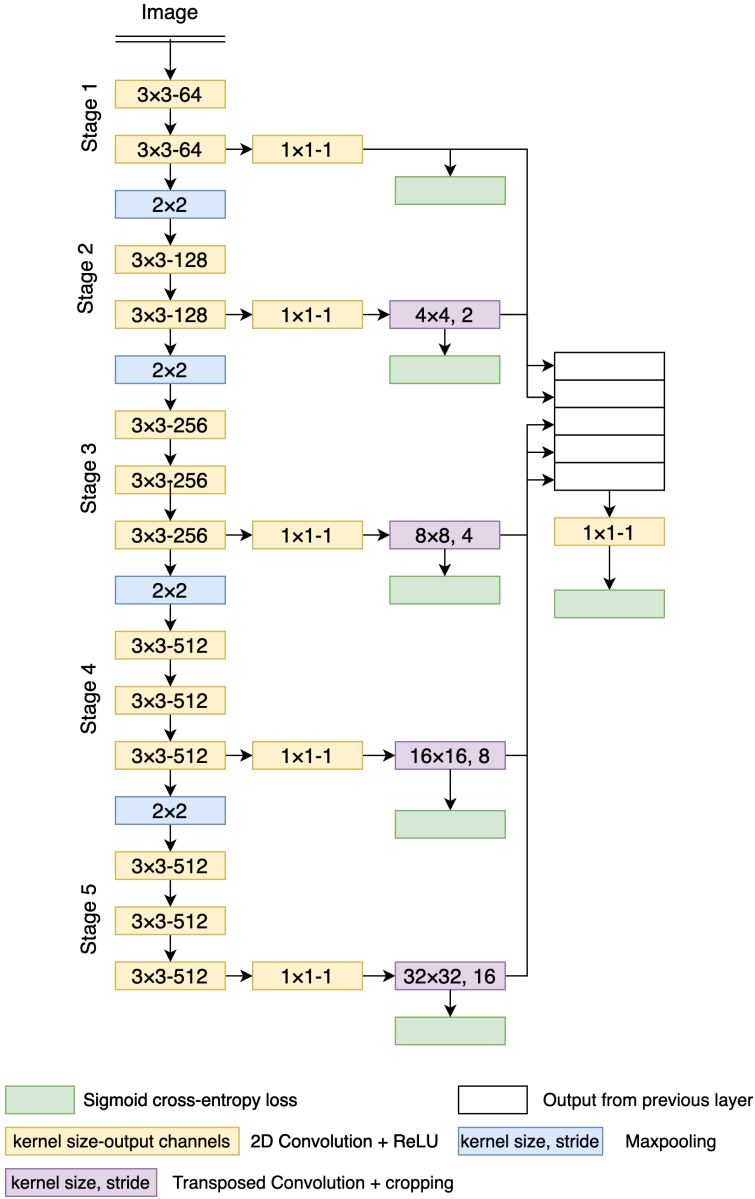

Figure 3: MS-CNN - The multi-scale architecture of Rahnemoonfar et al. (2021) which forms our base model

### 3.3.1. WaveNet

In Figure 3, from stage 2 to 5 of the base architecture, the feature maps get downsampled to a scale of 2x to 16x (with respect to the input image) to form the side outputs. Multi-level wavelet transforms of the input image also give us downsampled features at similar scales, which are sharp, denoised, and contain the horizontal, vertical, and diagonal detail coefficients (see subsection 3.1). By fusing wavelet information of the image to every scale, local as well as global, of the multi-scale network, the side outputs can be enriched for feature extraction. Hence, in this architecture, we take a level 4 wavelet transform (since there are four stages which have downsampled resolution as compared to the input image - stages 2 to 5) of the input image and fuse the detail coefficients in the following manner: we concatenate all three detail coefficients i.e. $H, V, D$ of a wavelet transform of level $l$ of the input radar echogram to the side output $l + 1$ of the base architecture. This means that the detail coefficients from level 1 of the wavelet transform fuse with side output 2 of the architecture, detail coefficients from level 2 of the wavelet transform fuse with side output 3 of the architecture, and so on. The WaveNet architecture is shown in Figure 4 .

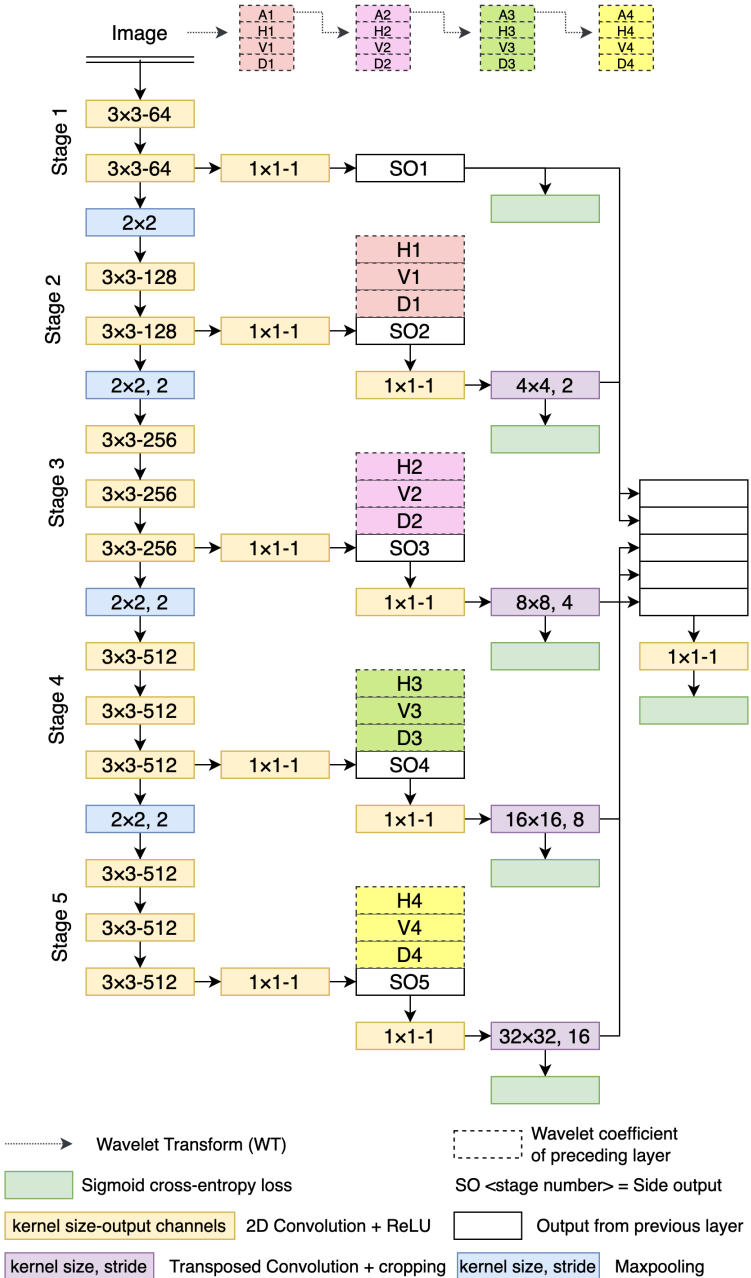

Figure 4: WaveNet - A wavelet based architecture. Here, the input image goes through a multi-level wavelet transform, where each level is shown with a different color, ranging from light pink to yellow.

### 3.3.2. Skip-WaveNet

The side outputs of MS-CNN are feature maps containing image patterns extracted at different scales, from local to global. These feature maps contain an improved representation of the features, i.e. are 'more learned', after every iteration of training and also contain noise in the form of unuseful network weights. Hence, they have a potential to be denoised further which can add value to overall network training. So for the Skip-WaveNet archiecture, we take a wavelet transform of each side output and add it as an extra layer to the successive side output to supplement network learning. This will help form skip connections, support residual learning (Bae et al., 2017; He et al., 2016) and also propagate information between scales. Since the wavelet transform is at a downsampled resolution to its input signal, the dimensions of a transform of side output $s$ will be the same as the dimensions of side output $s + 1$. Further, by fusing the wavelet transform of a scale to its successive scale, we are also propagating information between scales which can support network learning. With this intuition, we form the Skip-WaveNet architecture as follows: we take a level 1 wavelet transform of a side output $s$ ($s \in [1, 4]$) and fuse its detail coefficients to side output $s+1$. In this case, for a given input image, the wavelet transforms will be renewed after each epoch of training. This can be compared against the WaveNet architecture where the transforms generated for an input image will be the same during all epochs. Hence, such multi-scale 'dynamic' wavelets should improve the denoising and feature extraction capabilities of the network. The architecture of Skip-WaveNet is shown in Figure 5 .

Apart from the above defined modifications, WaveNet and Skip-WaveNet are trained in the same way as the base architecture MS-CNN, i.e. they are trained with deep supervision through binary cross entropy loss, and have the final fusion of five side outputs, etc.

### 3.4. Experimental Setup

All the architectures, including the base architecture, and wavelet combined neural networks share the same hyperparameters as those used in Xie and Tu (2015) and Rahnemoonfar et al. (2021). For our experiments, we augment the training dataset with scaling factors $\in [0.25, 0.5, 0.75]$ and a left-to-right flipping. The augmentation helps in expanding the training dataset to 6430 images, which we train for 15 epochs. All networks took approximately 10 hours, on an average, when trained on NVIDIA GeForce RTX 2080 Ti GPU with an Intel Core i9 processor.

### 3.5. Post-processing and Evaluation

We perform non-maximum suppression (NMS) on the network outputs to make finer predictions of the layers. NMS gives grayscale predictions, which we compare against our ground truth by evaluating the ODS (optimal dataset scale) and OIS (optimal image scale) F-scores (Y. Liu et al., 2019). The ODS F-score looks for a single threshold across the entire dataset which can binarize the predicted image, and give the most optimum F-score with respect to the ground truth. On the other hand, OIS F-score searches for a similar optimum threshold for each image. The optimum F-scores thus found from each image are then averaged over the entire dataset to obtain the OIS F-score. Here, F-score refers to the standard F1-score used in computer vision, defined as Equation 3.6. In this equation, TP, FP, and FN are True Positives, False Positives, and False Negatives, respectively. We also use average precision (AP), which is the area under a precision-recall curve (Su et al., 2015), as an evaluation metric of network performance.

$$F = \frac{TP}{TP + \frac{1}{2}(FP + FN)} \tag{3.6}$$

Based on the optimum threshold we get, we binarize the network outputs to get predictions similar to the ground truth. We then calculate the row indices (or the "range bin" indices of an echogram) of each layer in the predicted image matrix, and compare it with corresponding row indices of the manually labelled ground truth to evaluate the mean absolute error (MAE) of layer depth using Equation 3.7. The overall MAE was then obtained by estimating the MAE for each layer in every echogram for the entire test set using Equation 3.8.

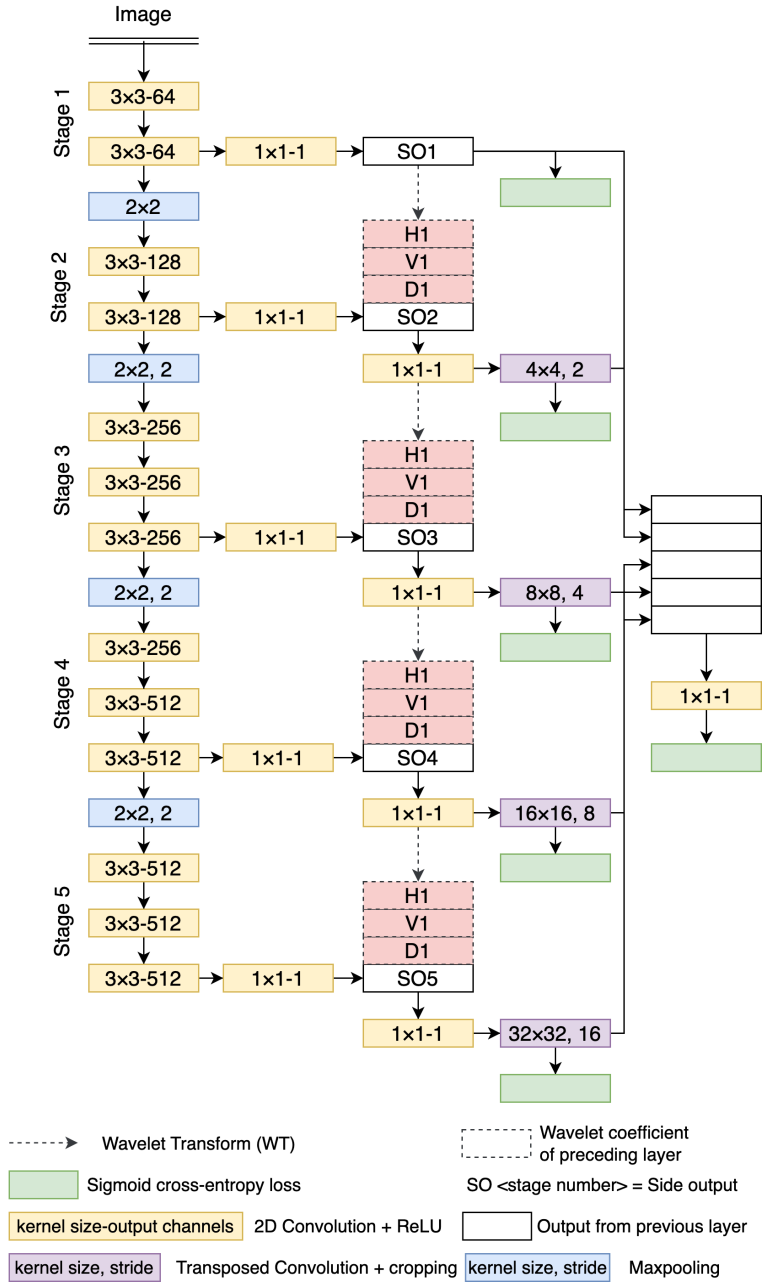

Figure 5: Skip-WaveNet - A wavelet based architecture with skip connections. Here, we take a level 1 wavelet transform of each side output. This is in contrast to WaveNet where the wavelet transform was of the input image.

$$MAE_j^k = \frac{\sum_{i=1}^{i=W_k} |GTD_i^{k,j} - PD_i^{k,j}|}{W_k} \tag{3.7}$$

$$MAE = \frac{1}{T} \sum_{k=1}^{k=T} \frac{\sum_{j=1}^{j=N_k} MAE_j^k}{N_k} \tag{3.8}$$

In Equation and 3.7 and 3.8, $GTD$, and $PD$ are the ground truth and predicted layer depths, respectively, in terms of number of pixels. For example, $PD_i^{k,j}$ represents the the row index of the $j^{th}$ predicted layer in the $i^{th}$ column of echogram $k$. $W_k$ and $N_k$ respectively represent the total number of columns (width), and the total number of traced layers, in echogram $k$. $T$ represents the total number of echograms in the test set.

### 3.6. *Comparison with In-Situ Stake Measurements*

The snow thickness and annual accumulation rates can be computed from traced annual layers. To assess the effects of tracing imperfections on the annual accumulation rate estimates, we compare the radar-derived accumulation rates with the ground measurements. Of the available in-situ stake measurements compiled by SUMup group (see Section 4), the closest measurement to the flight line of the radar data set used in this paper is chosen as the reference for comparison. The stake location is only ∼16 km away and is thus suitable for reliable comparison. Given the scale of ice-sheet, we do not expect the accumulation to vary at such a short distance. We estimate the radar-derived snow accumulation rates similar to J. Li et al. (2023) which uses Benson's density-depth model (Benson, 1960) and an interpretation model from Clarke et al. (1989). The in-situ stake measurements provide the required initial values of snow density and accumulation at the surface for the interpretation model (Clarke et al., 1989). The monthly stake-measured snow accumulations between September and August in two consecutive years are added up as the annual accumulation rate. The MAE effects are assessed by comparing the radar-derived accumulation rates with and without the MAE of each layer, which is calculated by Equation 3.8.

## 4. Dataset

We use Snow Radar images publicly hosted by the Center for Remote Sensing and Integrated Systems (CReSIS) (CReSIS, 2012). This dataset was captured over different regions of the GrIS and used to analyze the ice sheet's annual accumulation rates (Koenig et al., 2016). The blue line on the map in the left panel of Figure 6 shows one of the transects. The right panel of the figure presents the radar echogram of a 250-km section along the transect showing the snow layers to the depth of ∼ 12 m. This transect contains 1286 training images and 321 test images where each image has a vertical resolution of approximately 2.5 cm, and an along-track resolution of 14.5 m (Table 1). We use binary labels for training and evaluating our model, where the top of each snow layer is labelled as '1', and all other pixels are labelled as '0'. These labels are prepared manually by scientific experts tracing out the echograms through visual inspection.

To validate our model predictions with in-situ observations, we use the SUMup dataset (Dibb & Fahnestock, 2004; Thompson-Munson et al., 2022). This dataset contains surface mass-balance field observations of snow density, accumulation on land ice and its errors over ∼60 year time period across Arctic and Antarctic. This data set contains ice cores/snow pits, radar isochorones and stake measurements. In the stake measurements, midpoint depth, snow layer densities, monthly snow accumulation and error in accumulation are available. Figure 6 highlights the ice cores from SUMup dataset which are closest to our flight line in 2012. For fair comparison, we chose a stake (Dibb & Fahnestock, 2004) whose measurements have a temporal overlap with the radar data collection time-span.

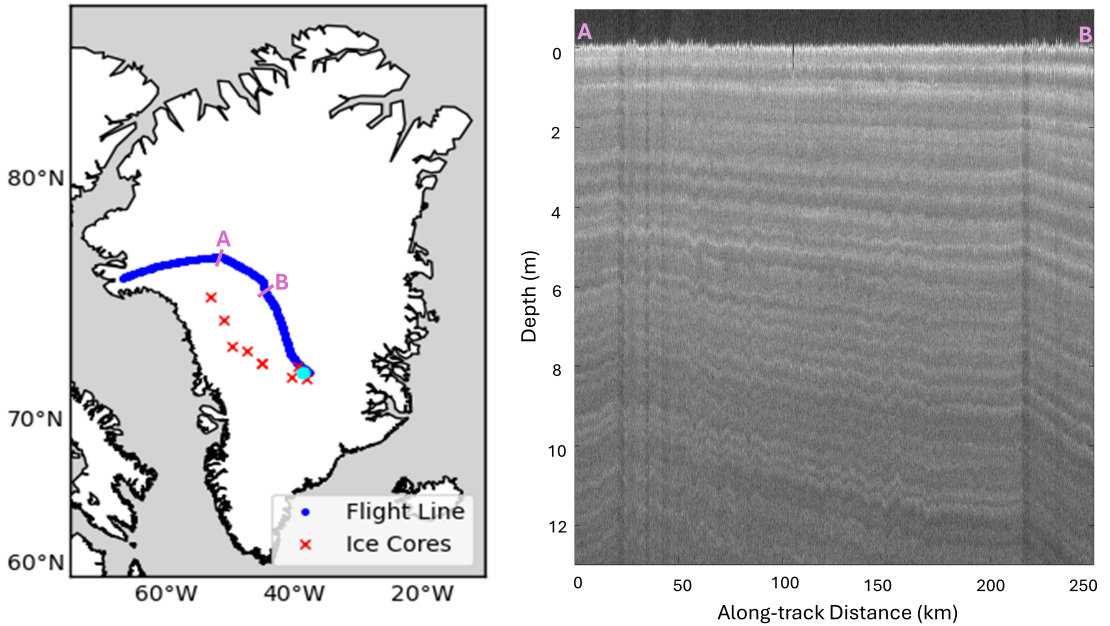

Figure 6: Left: Flightline of NASA Operation IceBridge 2012 in blue, nearest ice cores to the flightline in red, and the site which has a temporal overlap with IceBridge data in cyan. Right: A radar echogram spans 250 km along this flight line, marked by points 'A' and 'B'.

Table 1: Key parameters of the Snow Radar sensor used for data collection

| Parameter | Value |
|---|---|
| Bandwidth | 2-8 GHz |
| Pulse duration | 250 $\mu$ s |
| PRF | 2 kHz |
| Transmit power | 100 mW |
| Intermediate frequency range | 62.5 -125 MHz |
| Sampling frequency | 125 MHz |
| Range resolution | $\sim$ 4 cm |
| Along-track resolution | 14.5 m |

## 5. Results and Discussion

We note that training with all five side outputs of our base MS-CNN model on the new dry-zone dataset by CReSIS is detrimental to performance, since there are hardly any global features or contours that this network (or humans in general) can detect (see for example, the radar echograms in Figures 8 and 9). Hence, we use only the first four side outputs and the fuse layer to train the baseline MS-CNN model. This is not the case for wavelet combined neural networks which are able to detect layer-contours across all scales, and can generate the fifth side output as well, capturing global context. In this section, we showcase our results on wavelet combined neural networks vs the baseline model, and also discuss about the wavelet architectures WaveNet vs Skip-WaveNet. We also compare our performance with the only other work on Snow Radar layer tracing (Wang et al., 2021). Further, in Figure 7, we showcase Skip-WaveNet's ability in tracing snow layers along a 50-km transect.

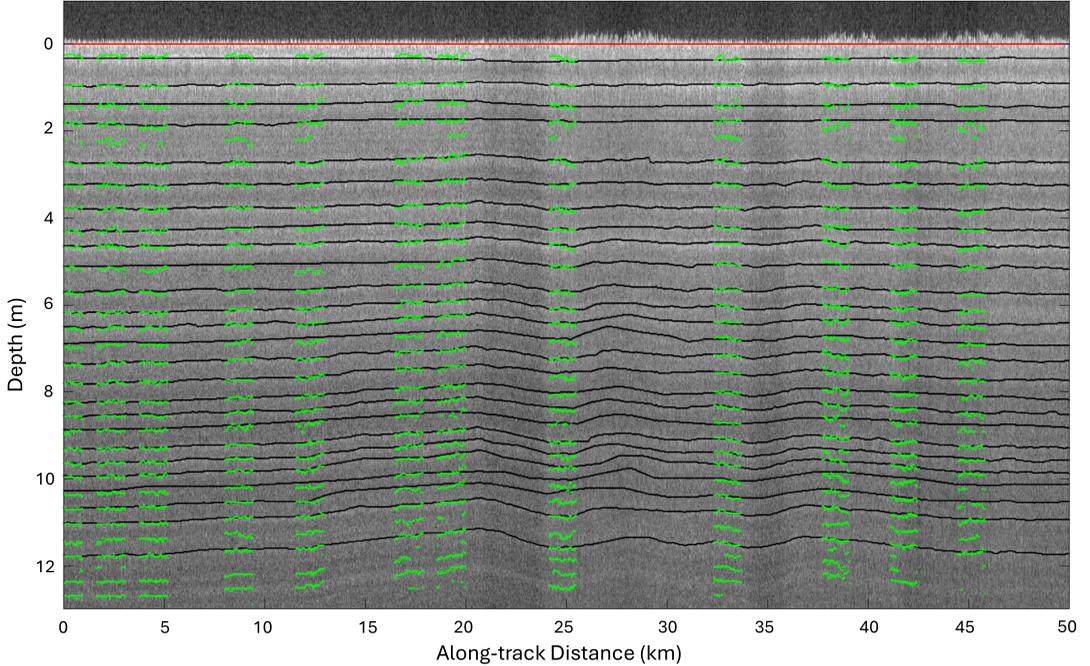

Figure 7: Radar echogram of a 50-km transect with the ground truth marked in black. Layers predicted by Skip-WaveNet on the test regions shown in green. The network is able to trace faint layers deeper than 12 m, whose ground truth is difficult to prepare.

Table 2: Evaluation metrics obtained across different model architectures on the test set. Highest scores highlighted in bold

| Network | Wavelet | ODS | OIS | AP | MAE |
|---|---|---|---|---|---|
| CNN3B+RNN | None | NA | NA | 0.853 | 8.730 |
| MS-CNN | None | 0.852 | 0.866 | 0.918 | 9.492 |
| WaveNet | haar | 0.876 | 0.888 | 0.936 | 3.517 |
| | db | 0.870 | 0.883 | 0.931 | 3.541 |
| | dmey | 0.835 | 0.851 | 0.905 | 3.967 |
| Skip-WaveNet | haar | 0.880 | 0.892 | 0.938 | 3.451 |
| | db | 0.879 | 0.890 | 0.937 | 3.438 |
| | dmey | **0.886** | **0.898** | **0.943** | **3.309** |

### 5.1. *Wavelet combined neural networks vs MS-CNN*

We tabulate the ODS and OIS F-scores, the average precision (AP), as well as the mean absolute error (MAE) of depth estimates in Table 2. These estimates are calculated on the final fuse layer of our experiments. From Table 2, we see that all wavelet combined neural networks, except for WaveNet with a dmey wavelet, perform better than the baseline MS-CNN model. The wavelet based networks also give higher AP and MAE scores (3.31 pixels) as compared to CNN3B+RNN (Wang et al., 2021). Wang et al. (2021) were able to predict the top four to six layers from an echogram, while the multi-scale architecture, whether MSCNN or wavelet-based, can trace out the deeper, twenty to thirty, layers from the echogram (Figures 8-10).

In the wavelet-based networks, Skip-WaveNet with a dmey wavelet achieves the highest F-scores, with all Skip-WaveNet models performing better than WaveNet. In the rest of the manuscript, we use the words Skip-WaveNet and Skip-WaveNet-dmey interchangeably. The wavelet transforms help in improving the ODS F-score by 2.11%-3.99% and the OIS F-score by 1.96%-3.7%, over the baseline architecture, for different experiments. Thus, the wavelet transforms not only help in denoising feature maps, but also extract features at the global scale (fifth stage), which the comparative MS-CNN does not. Since images are extremely noisy with different images having varying levels of intensity and contrast (see the different sample radar echograms in Figures 8-10), the layers can be almost impossible to see by human eyes. Hence, extracting features at a global fifth stage is a useful property that the wavelet architectures have achieved. In Figures 8-10, it can be seen that the layers detected by the wavelet network give a consistent and useful tracking result than the baseline MS-CNN model. The baseline model is misguided by a lot degraded features in the input radar echograms, and falsely classifies redundant backscatterings as a layer, as can be seen by its intermediary predictions in between two layers, highlighted in red in the figures. The layers predicted by MS-CNN mostly have missing values and are discontinuous in nature (i.e. have gaps in between) giving a reduced tracking accuracy (Table 2). MSCNN's MAE is worse by ~6 pixels which will give inaccurate thickness estimates eventually adding to uncertainties in accumulation rates (Kahle et al., 2021). Further, it should be noted that all multiscale architectures (MS-CNN, WaveNet, and Skip-WaveNet) have the same baseline CNN architecture, and hence the same number of trainable parameters. In WaveNet and Skip-WaveNet, the wavelet coefficients are additional filters computed from either the convolutional layers (side outputs) or the input images, and hence neither act as parameters which require training nor are updated by backpropagation. In this regard, both Skip-WaveNet and WaveNet are able to achieve a higher tracking accuracy even though their complexity, in terms of number of trainable parameters, is the same as that of MS-CNN.

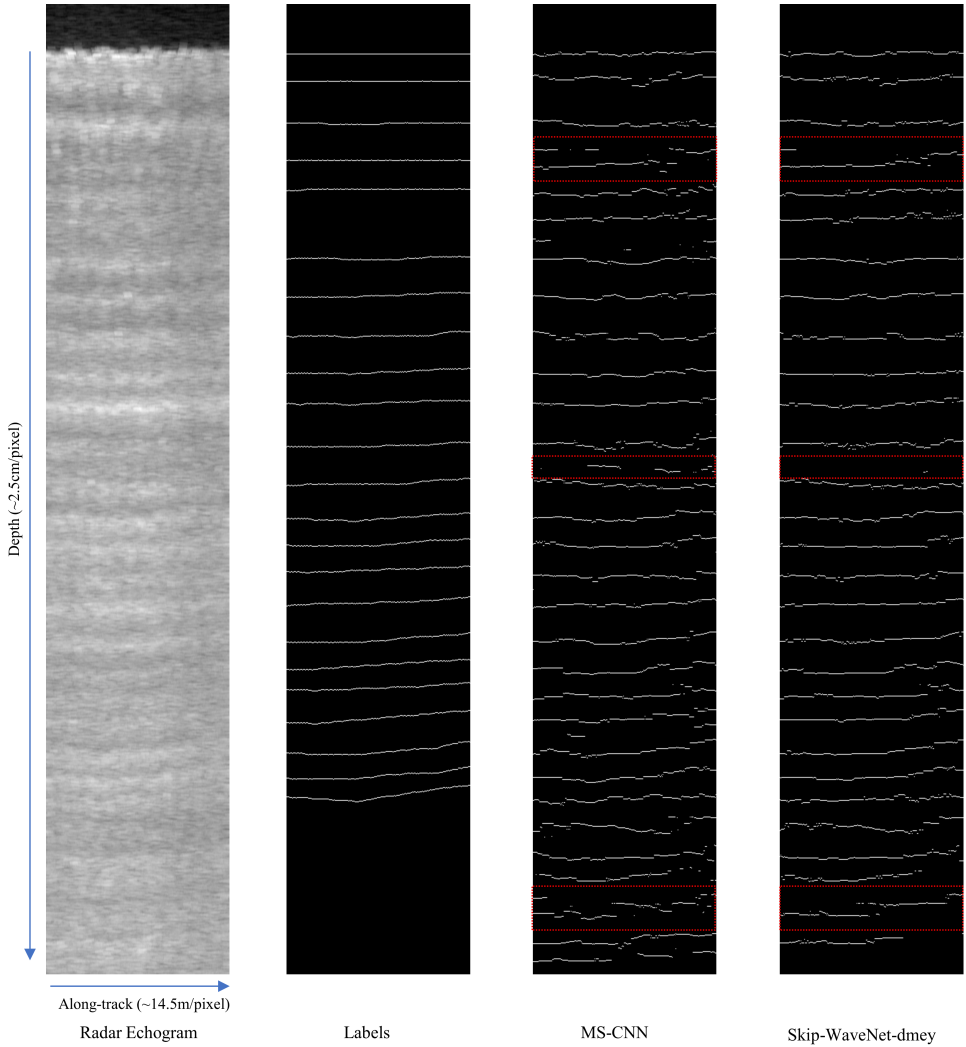

Figure 8: Qualitative comparison of model outputs. From left to right, are the Snow Radar echogram, the ground truth labels, output by MS-CNN (Rahnemoonfar et al., 2021) and by our proposed Skip-WaveNet-dmey model. The red boxes highlight some of the regions where Skip-WaveNet gives a cleaner prediction compared to MS-CNN. The two networks achieved an MAE of 3.74 and 4.22 pixels, respectively, across all the layers in this echogram.

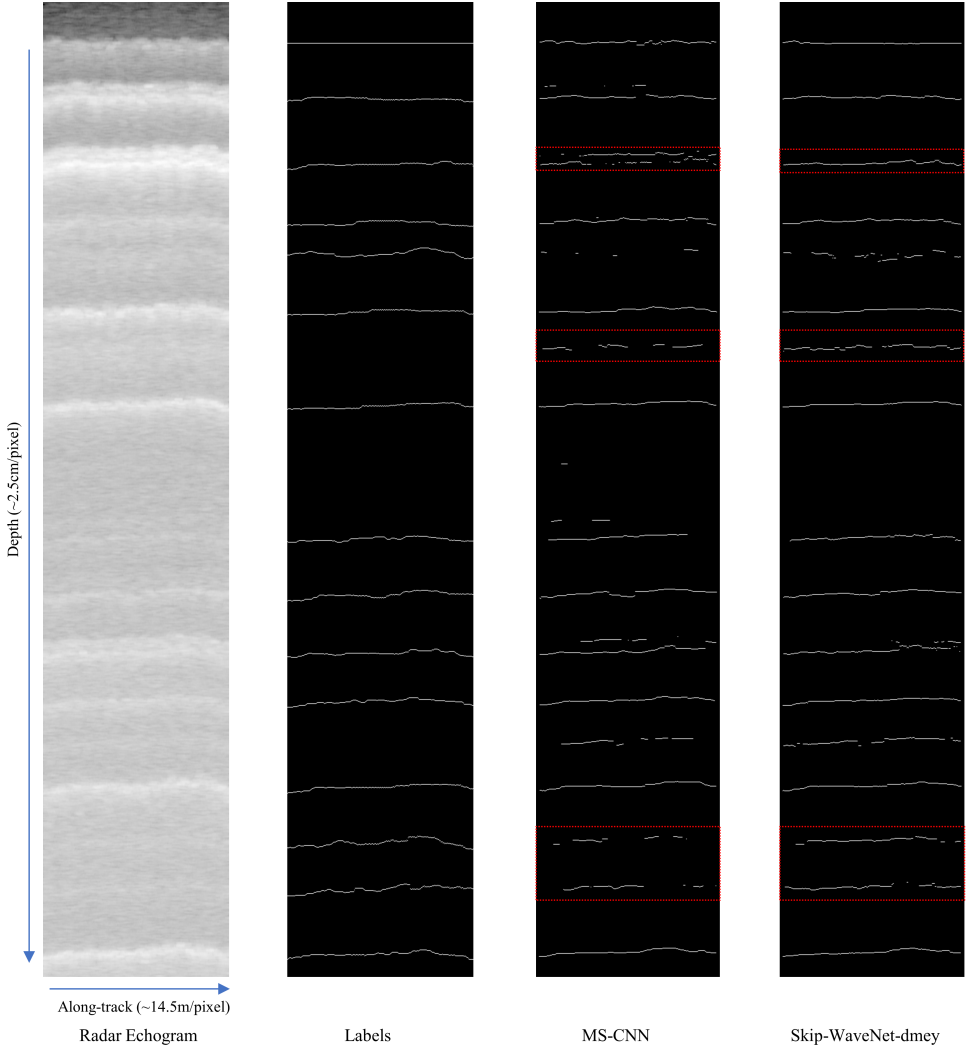

Figure 9: Qualitative comparison of model outputs. From left to right, are the Snow Radar echogram, the ground truth labels, output by MS-CNN (Rahnemoonfar et al., 2021) and by our proposed Skip-WaveNet-dmey model. The red boxes highlight some of the regions where Skip-WaveNet gives a cleaner prediction compared to MS-CNN. The two networks achieved an MAE of 3.17 and 4.22 pixels, respectively, across all the layers in this echogram.

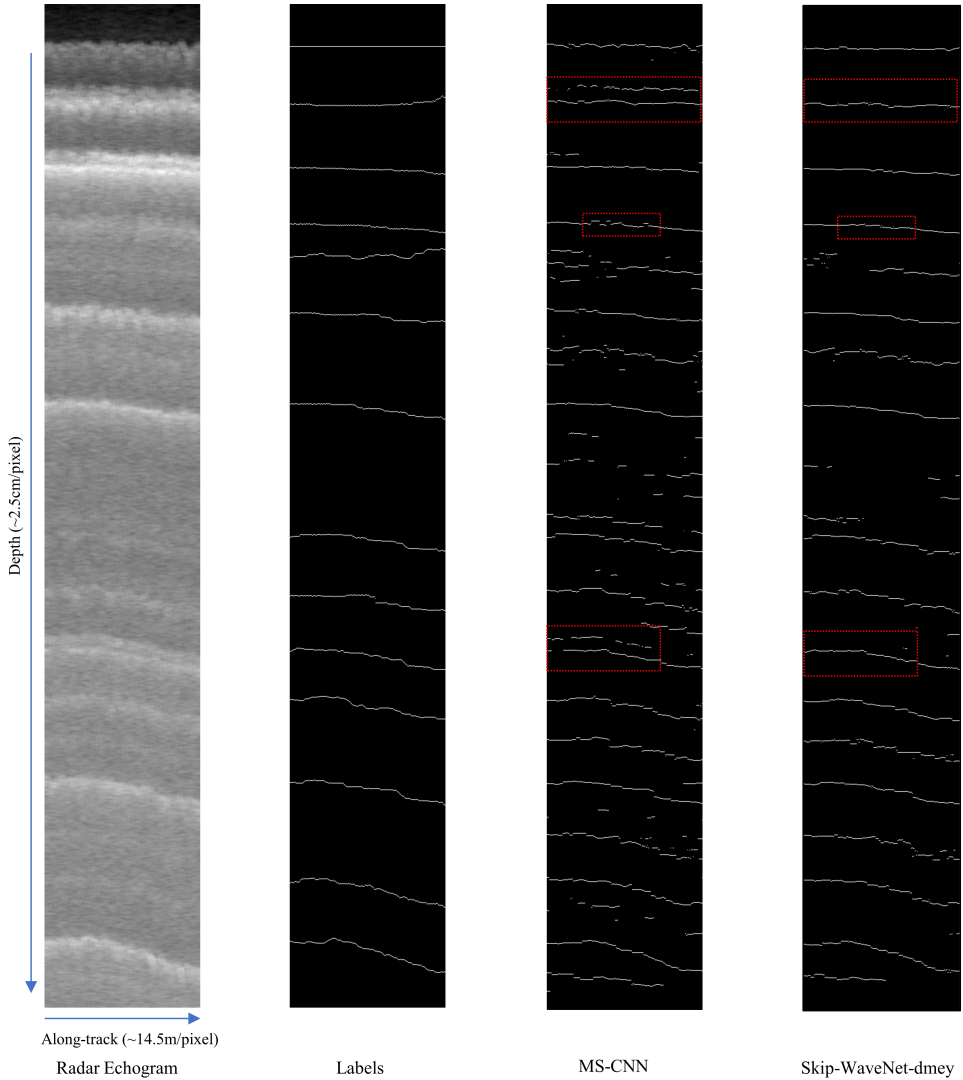

Radar Echogram          Labels          MS-CNN          Skip-WaveNet-dmey

Figure 10: Qualitative comparison of model outputs. From left to right, are the Snow Radar echogram, the ground truth labels, output by MS-CNN (Rahnemoonfar et al., 2021) and by our proposed Skip-WaveNet-dmey model. The red boxes highlight some of the regions where Skip-WaveNet gives a cleaner prediction compared to MS-CNN. The two networks achieved an MAE of 5.93 and 8.09 pixels, respectively, across all the layers in this echogram.

### 5.2. *Skip-WaveNet vs WaveNet*

Our experiments confirm that wavelet based architectures, by supplying additional information from the wavelet coefficients, enhance feature extraction. From Table 2 we see that Skip-WaveNet performs better than others, where the wavelet transform obtained in each epoch is from a learned and improved side output (so the wavelets are 'dynamic' and change after each epoch), whereas in WaveNet, the multi-level wavelet transforms are always fixed for a given input image for every epoch (the wavelets are in a way 'static' across the entire training period). We further see that the performance of the WaveNet experiment deteriorates with the wavelet complexity, whereas the performance of Skip-WaveNet improves with wavelet complexity (ODS F-score of dmey > db > haar). What would be interesting for future studies is to investigate is how the different types of wavelets are affecting network learning in different ways.

### 5.3. *Validation with Snow Stake Measurements*

In addition to the geographical proximity, the measurements at the selected snow stake site (Dibb & Fahnestock, 2004) (the cyan dot in the left panel of Figure 6) span the temporal range of 2003 to 2016, thereby exhibiting a temporal overlap with the radar data collection period. This temporal alignment renders it particularly well-suited for conducting comparative analyses with our model predictions applied to the 2012 IceBridge dataset. We use this snow stake site and find the test echogram closest to it, to compare Skip-WaveNet's predictions over this echogram and the measurements at the selected stake site. Figure 11 shows the annual accumulation rate for different years (2004-2012), as well as the associated error in these measurements at the stake measurement site. Similarly, the plot also shows the radar derived accumulation rate and its error due to the mean absolute error achieved by SkipWaveNet-dmey for the first eight annual snow layers between 2004 and 2011. The depth of the 2004 layer is ~4.5m below the surface. As can be seen from the plot, the estimates from SkipWaveNet are within the error bounds of the stake accumulation measurements. Moreover, the radar-derived accumulation rates from training labels (i.e. without MAE, and marked by *) agree well with the accumulation rate measured at the stake site. The MAE of the first eight layers in SkipWaveNet's predictions is 2.2 pixels corresponding to an error in accumulation rate estimation of ~0.011 m w.e. a-1. Therefore the tracing improvement of 6.2 pixels by Skip-WaveNet dmey over MS-CNN (see Table 2) results in reducing the uncertainty by 0.031 m w.e. a-1 in accumulation rate estimation. We also compare the radar-derived accumulation rates from the mean absolute error of SkipWaveNet and MSCNN for the same echogram in Figure 12 which shows that Skip-WaveNet was able to give a smaller error in accumulation rates estimates as compared to MS-CNN.

   At this stake site, however, the available density measurements were made between 1997 and 2002 in different months. In Figure 13 we plot these density measurements, made from the surface to a depth of 3 meters, in the form of blue circles. The red line shows the density-depth profile obtained from linear fitting with least mean square error. The orange line shows the density-depth profile used by the interpretation model (Clarke et al., 1989) for the accumulation rate estimation from tracked snow layers in radar echogram. Figure 14 presents the corresponding profiles of snow deposition age and dielectric constant versus depth, respectively.

### 6. Conclusion

Radar echograms are used for capturing the profile of multiple snow layers accumulated on top of an ice sheet. These echograms can have varying characteristics that can make snow layers difficult to detect visually or through automated algorithms. In this work, we use wavelet transforms to extract multi-scale textural information from radar echograms to supplement neural network learning. Our novel architecture helps in improving layer detection from noisy echograms. We also find that using wavelet transforms of intermediate side outputs and forming skip connections with them help in feature extraction and network learning, as compared to using 'static' wavelets of the input radar echogram. We show that our algorithm can be used for layer tracking and calculating snow thickness accurately, thus reducing accumulation rate error as compared to in-situ stake measurements. Since our models are data-driven in nature they are mainly effective for echograms captured in similar dry snow zone regions and can provide thickness estimates for glaciological models to use. Our future works are looking at tracing the layers for ablation zones (where the layers are much fainter, as seen through radar), leveraging larger datasets to incorporate regional variation and improve generalizability, as well as using multiple sensors to introduce complementary information.

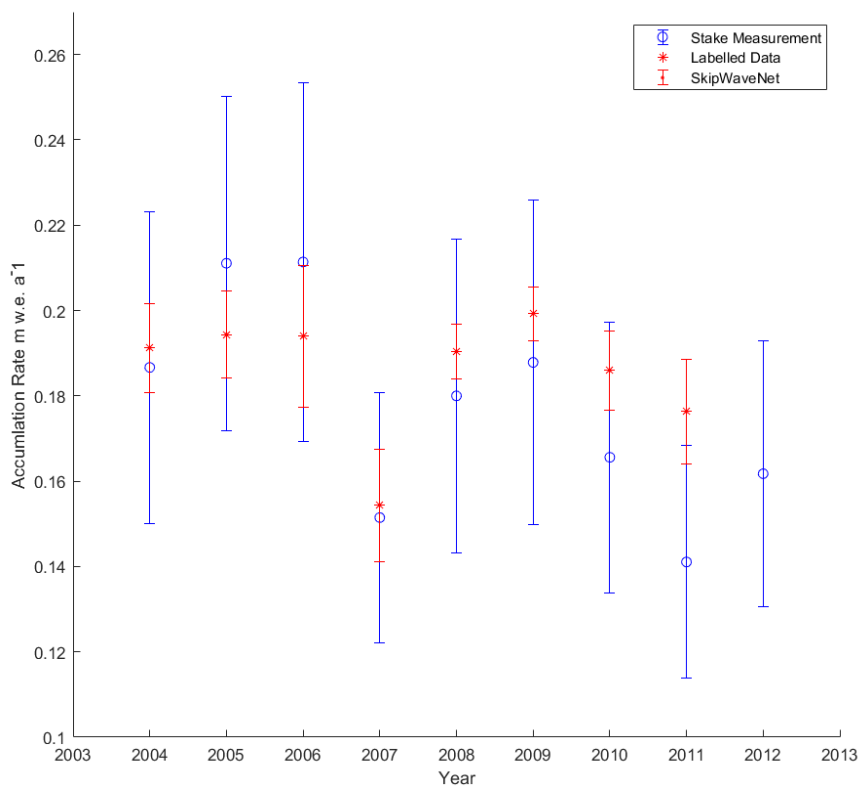

Figure 11: Comparison of the accumulation rate and its error at the site of stake measurements (blue) with the radar-derived accumulation rate from the traced annual layers marked by the training labels (*), and with the mean absolute error achieved by SkipWaveNet (red).

**Funding Statement.** This study was funded by NSF BIGDATA awards (IIS-1838230, IIS-1838024), the U.S. Army Grant No. W911NF21-20076, IBM, and Amazon.

**Competing Interests.** None

**Data Availability Statement.** The dataset is publicly hosted by the Center for Remote Sensing and Integrated Systems (CReSIS) at CReSIS (2012).

**Author Contributions.** Conceptualization: M.R. Resources: M.R; J.P; A.G. Software & Methodology: D.V; M.Y.; M. R; O.I; J. L. Data curation: O.I. Data visualisation: D.V. Formal Analysis: D.V; M.R. Snow accumlation rate comparison and validation: J. L; Supervision: M.R. Funding acquisition: M.R; Investigation: D.V; M.R. Writing original draft: D.V. All authors reviewed and approved the final submitted draft.

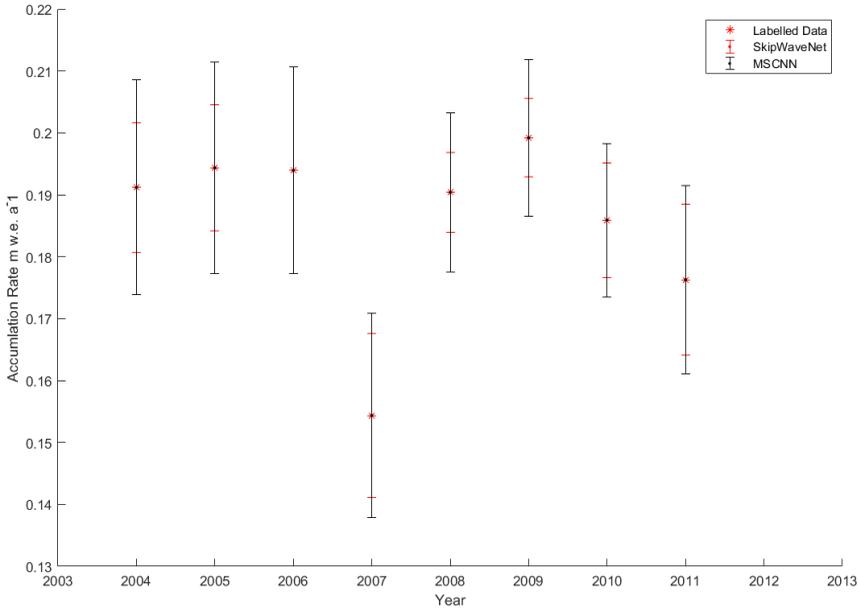

Figure 12: Comparison of the radar-derived accumulation rate from the mean absolute error of Skip-WaveNet (red) and MSCNN (black), both centered at the radar-derived accumulation rate from the traced annual layers (*). Clearly, SkipWaveNet gives a lower error.

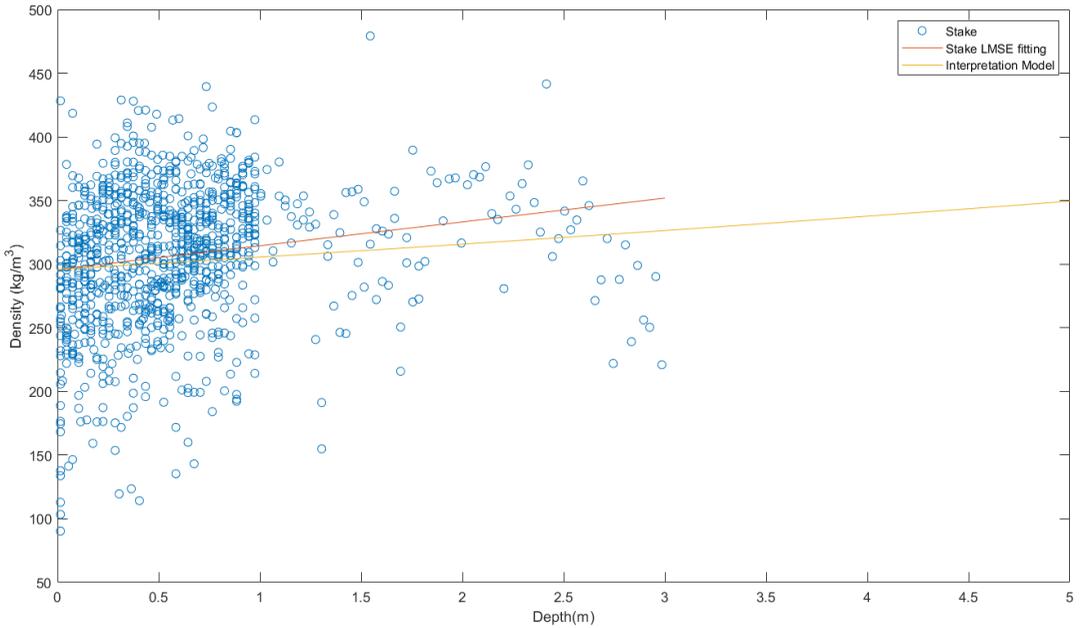

Figure 13: Snow density measurements at the stake site (Dibb & Fahnestock, 2004) in blue, their linear fit in red, and the density-depth profile (through the interpretation model of Clarke et al. (1989)) of snow layers traced from the closest radar echogram in orange.

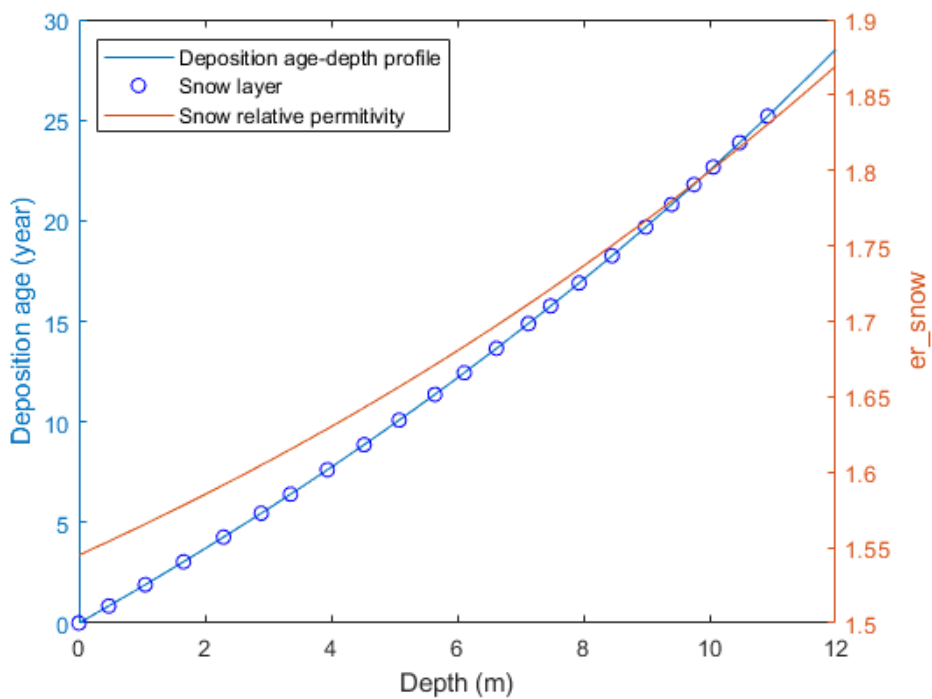

Figure 14: Profiles of snow layer deposition age (in blue) and dielectric constant (in red) versus depth estimated through the interpretation model of Clarke et al. (1989) at the location of the radar echogram closest to the stake site (Dibb & Fahnestock, 2004). Layers traced by Skip-WaveNet on this radar echogram are marked by circles.